\documentclass[10pt,twocolumn,letterpaper]{article}

\usepackage[pagenumbers]{cvpr} 
\usepackage{graphicx}
\usepackage{amsmath}
\usepackage{amssymb}
\usepackage{booktabs}
\usepackage{multirow}

\usepackage[pagebackref,breaklinks,colorlinks]{hyperref}

\usepackage[capitalize]{cleveref}
\crefname{section}{Sec.}{Secs.}
\Crefname{section}{Section}{Sections}
\Crefname{table}{Table}{Tables}
\crefname{table}{Tab.}{Tabs.}

\def\cvprPaperID{13650} 
\def\confName{CVPR}
\def\confYear{2024}

\begin{document}


\title{Learning from Observer Gaze: Zero-Shot Attention Prediction Oriented by Human-Object Interaction Recognition}

\author{Yuchen Zhou, Linkai Liu, Chao Gou\thanks{Corresponding author.}\\ 
 Sun Yat-sen University\\
{\tt\small \url{https://yuchen2199.github.io/Interactive-Gaze/}}  \\
{\tt\small \{zhouych37,liulk6\}@mail2.sysu.edu.cn, gouchao@mail.sysu.edu.cn} \\
}
\maketitle


\begin{abstract}
Most existing attention prediction research focuses on salient instances like humans and objects. However, the more complex interaction-oriented attention, arising from the comprehension of interactions between instances by human observers, remains largely unexplored. 
This is equally crucial for advancing human-machine interaction and human-centered artificial intelligence.
To bridge this gap, we first collect a novel gaze fixation dataset named \textit{IG}, comprising 530,000 fixation points across 740 diverse interaction categories, capturing visual attention during human observers' cognitive processes of interactions.
Subsequently, we introduce the zero-shot interaction-oriented attention prediction task (\textit{ZeroIA}), which challenges models to predict visual cues for interactions not encountered during training. 
Thirdly, we present the Interactive Attention model (\textit{IA}), designed to emulate human observers' cognitive processes to tackle the \textit{ZeroIA} problem.
Extensive experiments demonstrate that the proposed \textit{IA} outperforms other state-of-the-art approaches in both \textit{ZeroIA} and fully supervised settings.
Lastly, we endeavor to apply interaction-oriented attention to the interaction recognition task itself. Further experimental results demonstrate the promising potential to enhance the performance and interpretability of existing state-of-the-art HOI models by incorporating real human attention data from \textit{IG} and attention labels generated by \textit{IA}.
\end{abstract}

\section{Introduction}
\label{sec:intro}

\begin{figure}[t]
	\includegraphics[trim=22 15 15 15, clip=true, width=0.48\textwidth]{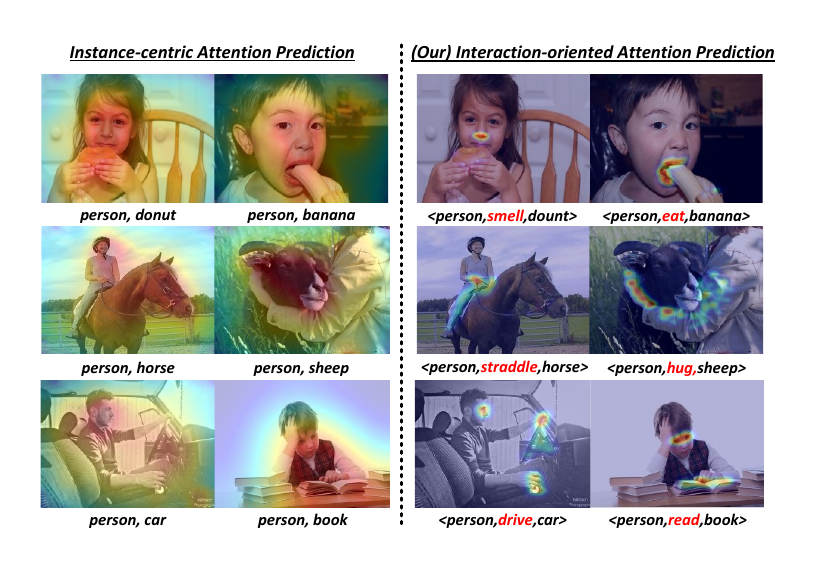}
	\caption{Previous attention prediction models have traditionally focused on the instance-level, primarily emphasizing foreground humans and objects. In contrast, our proposed interaction-oriented attention aims to capture subtler and more fine-grained visual cues associated with actions, such as body parts (row 1), human-object contact (row 2), and scene context (row 3). This proposition challenges the research community with a more intricate and cognitively demanding task.}
	\label{fig1}
\end{figure}

Human visual attention reveals a wealth of information about internal cognitive states, including bottom-up visual stimulation and top-down goal-directed empirical knowledge \cite{posner1990attention, zhang2020human, gou2017joint}. 
Predicting visual attention provides a valuable understanding of how humans perceive, comprehend, and interact with the world, especially within the context of goal-directed attention \cite{corbetta2002control,liu2021goal,shi2023top, pettine2023human, das2017human, chen2020air}. 
This holds profound implications for enhancing human-computer interaction \cite{katsini2020role,sen2020human,gou2023mal}, and contributing to fields such as assistance for visual impaired \cite{gurari2018vizwiz,gurari2020captioning}, education \cite{tafasca2023childplay,Chakraborty2023Predicting}, and autonomous driving \cite{gou2022cascade,Palazzi2017Predicting,10446034,zhou2024HKTSG}.


Goal-oriented attention prediction studies have made some progress recently \cite{yang2020predicting, mondal2023gazeformer, yang2022target, jiang2015salicon,fel2022harmonizing, yao2023teacher,attend2019,gou2022driver}, mostly focusing on predicting gaze fixation when searching or recognizing objects with relative explicit and invariant visual cues.
However, interactions \cite{zhou2023pit,liu2023learning,kim2021hotr, zou2021end, zhou2022toward}, another fundamental component of visual understanding that encapsulates how humans and objects interact with each other, remain largely unexplored.
Interactions pose greater cognitive challenges than objects, due to their diverse and subtle visual cues, as shown in Figure \ref{fig1}. 
Consequently, predicting interaction-oriented visual attention proves to be a more demanding task compared to previous efforts.
Moreover, the inherent diversity and almost infinite granularity of actions intersect with limited available data, necessitating the adoption of zero-shot learning. We name this problem \textit{Zero}-shot \textit{I}nteraction-oriented \textit{A}ttention prediction (\textit{ZeroIA}).

On the other hand, in the domain of human-object interaction (HOI) detection \cite{zhou2022human, liao2022gen, zhang2022efficient, zhang2022exploring, kim2023relational,zhou2023learning,tu2023agglomerative,kim2023relational, 10445997}, current state-of-the-art methods also face limitations in the interaction comprehension phase, even though they demonstrate proficiency in handling the object detection phase. This limitation is rooted in the inherent difficulty of capturing visual cues related to actions.


To alleviate these problems, we first introduce \textit{I}nteractive-\textit{G}aze (\textit{IG}), a novel dataset of gaze fixations capturing human cognitive process of interactions, filling a crucial gap in interaction-oriented visual attention research. IG contains 530,000 fixation points from 32 human observers, spanning 740 interaction categories, 80 objects, and 132 actions. 
Human observers are invited to capture key visual cues in various interaction scenes, while their visual attention during the cognitive process is recorded. 
All interaction scenes are selected from HOI benchmark datasets, HICO-det \cite{chao2018learning} and VCOCO \cite{gupta2015visual}. This means that IG has the potential to bridge the fields of visual attention and HOI detection and be a catalyst for advancing both research areas simultaneously.

Secondly, inspired by the cognitive processes of human observers, we further propose a goal-oriented attention prediction approach, termed \textit{I}nteractive \textit{A}ttention (\textit{IA}). 
Our approach begins by a set of clever interaction-oriented prompts and adapters to activate and leverage CLIP's powerful knowledge representation capabilities \cite{radford2021learning}, thereby facilitating zero-shot learning. This process involves establishing adaptive knowledge prototypes based on each visual scene, avoiding direct retrieval of fixed knowledge from CLIP, which may exhibit strong noun bias \cite{Momeni_2023_ICCV,HendricksN21,park2022exposing}. Guided by these knowledge, IA first focuses on the perception and understanding of individual instances, i.e., humans and objects. Then, it further understands the interactions unfolding between instances, ultimately completing the generation of interaction-oriented attention. Extensive experiments demonstrate that IA outperforms state-of-the-art approaches in both ZeroIA and fully supervised settings.

Furthermore, we explore how goal-oriented attention can be fed back to the goal itself, particularly establishing an initial bidirectional pathway connecting goal-oriented attention and action understanding. We introduce a general and effective HOI training strategy. This can supplement most state-of-the-art model loss with additional supervision for interaction-oriented attention. Remarkably, extensive experiments reveal the substantial potential of visual attention for HOI models from two aspects. (1) Aligning a limited yet valuable genuine human interaction-oriented attention from the proposed IG enhances the performance and interpretability of existing state-of-the-art HOI models.
(2) Integrating interaction-oriented attention generated by the proposed IA model further enhances HOI models, surpassing even the performance achieved in the context of human observers’ attention.
To summarize, the contributions of our work are three-fold:

\begin{itemize}
    \item Firstly, we address a crucial gap in interaction-oriented attention studies by introducing the ZeroIA problem and IG dataset, marking a pioneering endeavor as the inaugural gaze fixation dataset specifically designed for interaction analysis. Given the composite nature of IG, it has significant potential to facilitate the intersection of multiple domains, such as goal-directed attention and interaction comprehension.
    \item Secondly, we present the IA model, a novel approach crafted to emulate human cognitive processes for predicting high-quality interaction-oriented attention. Extensive experiments demonstrate IA outperforms other state-of-the-art attention prediction methods in both the ZeroIA and fully-supervised settings.
    \item Thirdly, we introduce a general and simple HOI training strategy with attention and demonstrate that both aligning a small yet precious genuine human attention from IG and aligning attention generated by IA can enhance the performance and interpretability of existing state-of-the-art HOI models. 
\end{itemize}



\section{Interactive Gaze (IG) Dataset}
\begin{figure}[t]
	\includegraphics[trim=0 0 0 0, clip=true, width=0.45\textwidth]{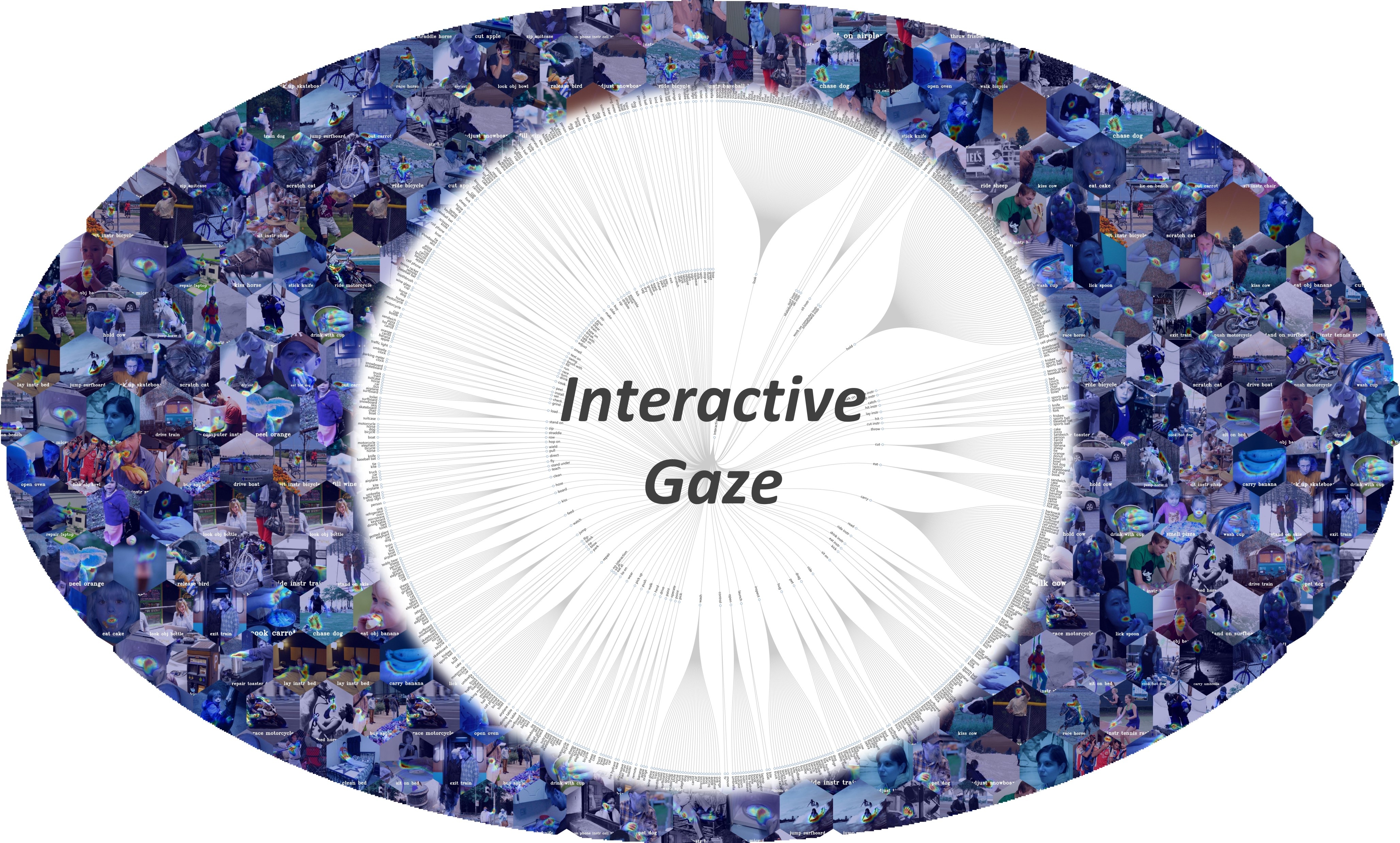}
	\caption{Our proposed IG is the first interaction-centric gaze fixation dataset, comprising 530K fixation points across 740 interaction categories.}
	\label{figIG}
\end{figure}

To study human visual attention oriented by HOI, we collect the Interactive Gaze (IG), the first large-scale interaction-centric gaze fixation dataset, as shown in Figure \ref{figIG}. 
IG comprises 6,299 interaction scenarios across 740 interaction categories, 80 object categories, and 132 action categories. IG captures the visual attention of 32 human observers during the cognition of these interaction scenarios, resulting in 530,000 corresponding fixation points.
IG holds substantial potential to bridge the domains of visual attention and action understanding, serving as a catalyst to jointly promote these two areas of study. We use mouse clicks to simulate gaze fixation for visual attention studies \cite{jiang2015salicon,linsley2017visual,attend2019,fel2022harmonizing}, and more details about the IG dataset are presented in the supplementary materials.

\textbf{Dataset Comparison.} Previous gaze fixation datasets have primarily concentrated on instances (e.g., objects, humans, animals, texts) as targets for recognition, search, and free-viewing, as illustrated in Table \ref{tab1}. 
In contrast, IG stands as the pioneering interaction-centric gaze fixation dataset. Given the inherent diversity, composability, and context dependency of interactions, interaction-oriented attention is associated with elevated cognitive challenges and empirical knowledge.

\begin{table}[t]
	\centering
	\caption{Comparison of fixation datasets. Previous work all focus on instances of objects, humans, animals, texts, etc.}
	\resizebox{0.45\textwidth}{!}{
		\begin{tabular}{lcccr}
			\toprule
			Dataset & Focus & Scene & Class & Fixation  \\ 
			\midrule
			SALICON \cite{jiang2015salicon} & Object & 10000 & - & 4600K  \\ 
			POET \cite{papadopoulos2014training} & Object & 6270 & 10 & 178K  \\ 
			People900 \cite{ehinger2009modelling} & Human & 912 & 1 &  55K  \\ 
			MCS \cite{zelinsky2019benchmarking} & Object & 2183 & 2 &  16K   \\ 
			PET \cite{gilani2015pet} & Animal & 4135 & 6 &  30K  \\ 
			COCO-Search18 \cite{yang2020predicting} & Object & 6202 & 18 & 300K  \\
			SalECI \cite{Jiang_2022_CVPR} & Object \& Text & 972 & 13 & 257K  \\ \midrule
			Our (IG) & Action & 6299 & 740 & 530K \\ \bottomrule
		\end{tabular}
	}
	\label{tab1}
\end{table}

\textbf{Data Selection \& Subsetting.} 
To bring the visual attention and HOI detection domains closer, we select image samples from HOI benchmark dataset VCOCO \cite{gupta2015visual} and HICO-det \cite{chao2018learning} as experimental materials. VCOCO consists of 24 action classes and 214 HOI classes, while HICO-det is larger and more challenging, covering 117 action classes and 600 HOI classes. We select 4,475 samples from the V-COCO training set (13 samples per HOI class on average) to train the attention prediction model. Additionally, we use 1,104 samples from the HICO-det test set and 720 samples from the V-COCO test set for testing the attention prediction model under ZeroIA and fully supervised interaction-oriented attention prediction tasks, respectively.





\section{Method}

\subsection{Problem Definition}
Given a sample of HOI pair, the goal of zero-shot interaction-oriented human visual attention prediction (\textit{ZeroIA}) is to generate an attention heatmap that reflects key visual cues for comprehending the interaction between human and object without any training for this interaction, as shown on the left of Figure \ref{fig3}. 
A complete HOI sample consists of five components: an RGB image $x \in \mathbb{R} ^{H\times W \times 3 } $, a bounding box of human $b^H$, a bounding box of object $b^O$, an object label $C_O$, and an interaction label $C_I$. The output of \textit{ZeroIA} is expressed as a spatial attention heatmap $m_{IA} \in \mathbb{R} ^{H\times W \times 1 }$.

\subsection{Interactive Attention}

\begin{figure*}[t]
	\includegraphics[trim=15 20 25 15, clip=true, width=1\textwidth]{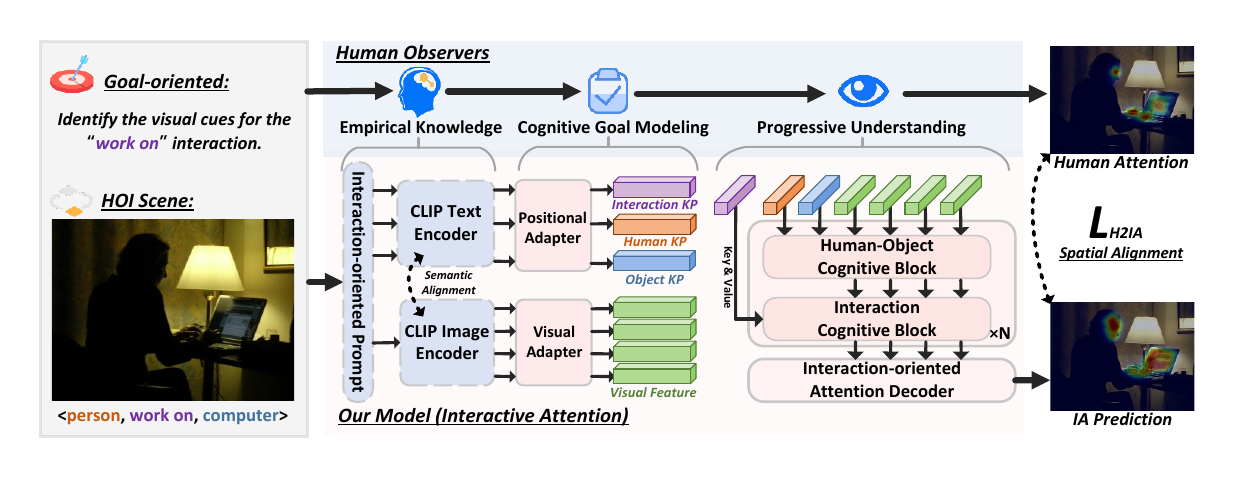}
	\caption{The overall architecture of our \textit{Interactive Attention (IA)}. Inspired by the HOI cognitive process of human observers, \textit{IA} is divided into three phases: empirical knowledge representation, cognitive goal modeling, and progressive understanding. First, a set of interaction-oriented prompts activate and leverage the robust knowledge representation capability of CLIP. Secondly, positional and visual adapters are introduced to acquire scene-adaptive human, object, and interaction Knowledge Prototypes (KPs) along with visual features of the HOI scene. Thirdly, guided by these KPs, \textit{IA} progressively comprehends the scene, starting with an instance-level understanding of humans and objects and deepening insight into their interactions. The decoder generates predicted attention maps, supervised by the real attention maps of human observers using $L_{H2IA}$.}
	\label{fig3}
\end{figure*}

Under goal-oriented cognitive tasks, humans tend to use their existing empirical knowledge and memory to model cognitive goals, and then progressively perceive and understand visual scenes with the goal. Inspired by the human cognitive process, we propose the \textit{I}nteractive \textit{A}ttention (\textit{IA}) to tackle the ZeroIA problem, as shown in Figure \ref{fig3}. 

In IA, we first activate the strong knowledge representation capabilities of CLIP \cite{radford2021learning} through a clever interaction-oriented prompt block (PB). Secondly, we design two tiny and learnable adapters to model the HOI cognitive task. It generates scene-adaptive knowledge prototypes (KPs) for humans, objects, interactions, and robust visual features. Guided by these knowledge, we progressively perceive and understand the HOI scene. Specifically, we first perceive the human and object through a human-object cognitive block (HOCB), then delve deeper into understanding the interaction that occurs between the human and objects through an interaction cognitive block (ICB), and finally decode the interaction-oriented attention map.


\textbf{Interaction-oriented Prompt.} Given an HOI sample, the interaction-oriented prompt block (PB) provides the visual-linguistic large model CLIP with clever inputs applicable to the HOI cognitive task to activate CLIP's powerful representation capabilities.
Due to the natural composability of HOI, PB considers humans, objects, and interactions separately, i.e., constructing human-centric, object-centric, and interaction-centric prompts and extracting their textual representations, respectively.

PB takes \textit{``person''} and object textual label $C_O$ as inputs to CLIP text encoder and extracts the human- and object-centric textual representations, $T_H$, and $T_O$. The human and object texts are merged into the interaction-centric prompt, i.e., \textit{``a photo of a person [$C_I$] [$C_O$]"} since they are closely related to the interaction. Then, the interaction-centric textual representation $T_I$ is extracted.
Correspondingly, CLIP's image encoder is leveraged to extract robust visual features $V = [v_1, ..., v_M]$ to take advantage of the semantic alignment between image and linguistics.

\textbf{Positional \& Visual Adapter.}
The features extracted from CLIP based on PB are somewhat applicable to the cognitive goal, but this is a fixed holistic representation modeling of the goal, which lacks the ability to adapt to each specific HOI sample, especially to diverse spatial configurations and visual scenes. As a result, we design the tiny positional adapter and visual adapter to provide HOI sample-aware knowledge that enables better modeling of HOI cognitive tasks in each specific sample.

Positional adapter (PA) provides sample-aware spatial information on textual representations:
\begin{equation}
K_H  = MLP([(T_H),Fourier(b^H)]),
\end{equation}
\begin{equation}
K_O  = MLP([(T_O),Fourier(b^O)]),
\end{equation}
\begin{equation}
K_I  = MLP([(T_I),Fourier(C_H),Fourier(C_O)]) ,
\end{equation}
where $K_H$, $K_O$, and $K_I$ represent adaptive knowledge prototypes of human, object, and interaction, $Fourier$ is the Fourier embedding, $MLP$ is a multi-layer perceptron, and $[\cdot ]$ is the concatenate operation. Visual adapter (VA) is implemented as two Transformer encoder layers to transform ${V}$ into refined visual features ${V}' = [{v_1}', ..., {v_M}']$.

\textbf{Human-Object Cognitive Block.}
The HOI scene is further learned under the guidance of knowledge prototypes of human and object ($K_H$ and $K_O$).
The self-attention mechanism with gating is introduced to establish cognitive dependencies between knowledge prototypes and visual features, then the human-object-aware visual feature $V_{HO}$ is obtained:
\begin{align}
V_{HO} = {V}' + Gate(SelfAttn([K_H, K_O, {V}'])),
\end{align}
where $Gate(\cdot)$ is a token gating operation that considers visual tokens only, $SelfAttn(\cdot)$ denotes a self-attention computation.

\textbf{Interaction Cognitive Block.} Expanding upon the instance-level cognition, including the human and object, the model further delves into comprehending higher-level interactions that occur between the human and object. To achieve this, the cross-attention mechanism with gating is introduced to guide the learning of more subtle interaction cues by interaction knowledge prototype and obtaining the primary HOI-aware visual feature $V_{HOI}$:
\begin{align}
V_{HOI} = V_{HO} + Gate(CrossAttn(V_{HO}, K_I)),
\end{align}
where we denote $CrossAttn(q,kv)$ as a cross-attention operation. After that, the general self-attention operation is conducted on $V_{HOI}$ to obtain the full HOI-aware visual feature ${V}'_{HOI}$:
\begin{align}
{V}'_{HOI} = V_{HOI} + SelfAttn(V_{HOI}).
\end{align}
\textbf{Interaction-oriented Attention Decoder.}
The goal of the decoder is to generate interaction-oriented attention heatmap $m_{IA} \in \mathbb{R} ^{H\times W \times 1 }$, we first reshape the full interaction-aware visual feature, ${V}'_{HOI}$, from a 2D shape of $\frac{HW}{p^2}\times C$ to a standard 3D feature map $\frac{H}{p}\times \frac{W}{p} \times C$, $C$ is the feature channel size, $p$ is the image patch size. Then, we adopt tiny 2-layers of $1 \times 1$ convolution with batch norm and ReLU activation to reduce the feature dimension to $\frac{H}{p}\times \frac{W}{p} \times 1$. Last, we adopt a bilinearly upsample operation to achieve the full image size $H\times W$.

\textbf{Loss Function.} We utilize the binary cross-entropy as a loss function for attention alignment, which can be expressed as follows:
\begin{equation}
\text{BCE}(p, y) = -\frac{1}{N} \sum_{i=1}^{N} \left( y_i \cdot \log(p_i) + \\
(1 - y_i) \cdot \log(1 - p_i) \right),
\end{equation}
\begin{equation}
L_{H2IA} = \text{BCE}(m_{H}, m_{IA}),
\end{equation}

where $p$ is the predictive probability, $y$ is the target, $N$ is the number of pixels in $m_{IA}$, $m_{H}$ is human observer's real attention heatmap.

\subsection{Training HOI Models with Interaction-oriented Attention}
\begin{figure}[t]
	\includegraphics[trim=22 15 15 15, clip=true, width=0.48\textwidth]{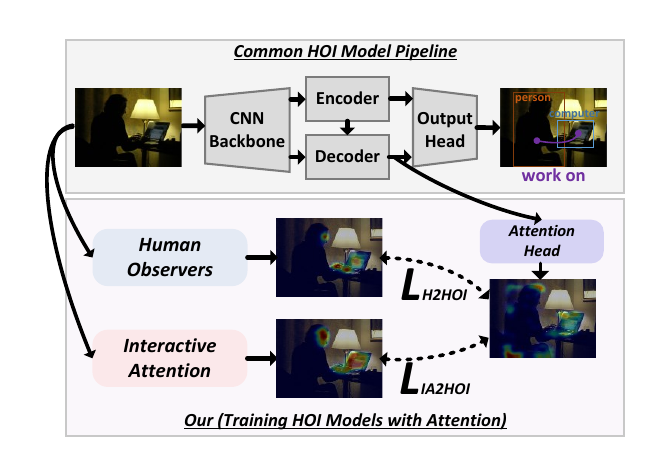}
	\caption{We incorporate aligned attention into the existing HOI model training pipeline, divided into two strategies: supervising by a limited amount of real attention from human observers and supervising by a large number of attention pseudo-labels generated by our proposed \textit{IA}.}
	\label{fig4}
\end{figure}

Here, we formulate the majority of recent state-of-the-art HOI models in the pipeline of CNN Backbone $\to$ Encoder $\to$ Decoder $\to$ Output Head. We enhance these HOI models by incorporating a supervision of interaction-oriented attention into the training process. This involves two strategies: supervising with a limited amount of real attention samples from human observers and supervising with a large number of pseudo-labeled attention samples from the IA model, as shown in Figure \ref{fig4}.

For these HOI models, we first introduce a simple attention head to extract and reshape the averaged attention maps $m_{HOI}$ from the last layer of cross-attention computation in the decoder. 
The attention head is intentionally kept simple for a more direct back propagation of guidance to the representation.
Subsequently, the corresponding human observer's attention heatmap $m_{H}$ and IA-generated pseudo-labels $m_{IA}$ are scaled down to match the size of $m_{HOI}$ through adaptive 2D max-pooling. Finally, $m_{H}$ and $m_{IA}$ are used to supervise $m_{HOI}$ for two kinds of strategies, respectively.

For the strategy of supervising by human observers, the attention alignment loss function $L_{H2HOI}$ and the overall HOI model's loss function $L$ can be expressed as 
\begin{equation}
L_{H2HOI} = \text{BCE}(m_{HOI}, m_{H}),
\end{equation}
\begin{equation}
L = \lambda_{1}L_{raw} + \lambda_{2}L_{H2HOI},
\end{equation}
where $L_{raw}$ represents the loss function of the original HOI model, which varies depending on the specific model, while $\lambda_{1}$ and $\lambda_{2}$ denote loss scaling factors.
Similarly, for the strategy of supervising by \textit{IA}, $L_{IA2HOI}$ and the overall loss function $L$ can be expressed as 
\begin{equation}
L_{IA2HOI} = \text{BCE}(m_{HOI}, m_{IA}),
\end{equation}
\begin{equation}
L = \lambda_{1}L_{raw} + \lambda_{2}L_{IA2HOI}.
\end{equation}


\section{Experiments}
\subsection{Implementation Details}
\textbf{Interactive Attention.} 
We apply 2 versions of CLIP (ViT-B/16 and ViT-L/14) to build the IA models, resulting in IA-B and IA-L, respectively. The output dimension of Fourier embedding is 64.
During training, we use AdamW as the optimizer with an initial learning rate of $1\times 10^{-4}$. The IA model is trained for 80 epochs with a learning rate reduction pre-20 epochs by a factor of 10. The momentum and the weight decay are set as 0.9 and $1\times 10^{-4}$. The experiments are conducted on four NVIDIA GeForce RTX 3090 GPUs with a batch size of 16.

\textbf{Training HOI Models with Attention Alignment.}
We apply our attention-guided training strategy on three typical state-of-the-art HOI methods: MUREN \cite{kim2023relational}, representing the one-stage structure, and UPT \cite{zhang2022efficient} and STIP \cite{zhang2022exploring}, representing the two-stage structure. In MUREN, we align the cross-attention map of the last layer in its interaction branch module with our interaction-oriented attention. In UPT, we first add two transformer decoder layers at the end of its interaction head, conducting cross-attention computation with global visual features, and align the cross-attention map of last layer. In STIP, we align the cross-attention map of the last layer in its structure-aware transformer module with our interaction-oriented attention. 
Due to variations in the raw losses of different methods, particularly for one-stage structure requiring additional supervision of object detection results, corresponding loss scaling factors differ. $\lambda_{1}$ and $\lambda_{2}$ are set to $1$ and $10$ in MUREN. $\lambda_{1}$ and $\lambda_{2}$ are set to $1$ and $6$ in UPT and STIP. Note that the IA model, responsible for generating interaction-oriented attention, is trained without exposure to samples from VCOCO test set and HICO-det test set, preventing any possibility of label leakage.

\subsection{Comparison methods \& Metrics.} 
We evaluate the performance of our \textit{IA} model on interaction-oriented attention prediction, comparing it with 10 state-of-the-art methods: ITTI \cite{itti1998model}, GBVS \cite{harel2006graph}, DeepGaze I \cite{KummererTB14}, DeepGaze IIE \cite{Linardos_2021_ICCV}, UMB \cite{Zhu_2023_ICCV}, ConvNext \cite{liu2022convnet}, MLNet \cite{dodge2018visual}, SSwin Transformer \cite{Jiang_2022_CVPR}, CLIP-ViT-B/16 \cite{radford2021learning}, CLIP-ViT-L/14 \cite{radford2021learning}.
Among these, ITTI and GBVS excel in early saliency prediction based on manual features, while the remaining methods are recent learning-based approaches. DeepGaze I and DeepGaze IIE utilize official pre-trained models for predictions, and UMB, ConvNext, MLNet, and SSwin Transformer are all trained with IG datasets. Additionally, we compute the similarity maps of two commonly used raw CLIP versions (CLIP-ViT-B/16 and CLIP-ViT-L/14). These similarity maps compute the similarity distance between image features and text features, serving as the raw CLIP's class attention maps \cite{li2023clip}.

We apply 4 metrics to measure performance: KL divergence (KLdiv), correlation coefficient (CC), similarity (SIM), and the area under the receiver operating characteristic curve (AUC). KLdiv, CC, and SIM are distribution-based metrics, while AUC is a localization-based metric. Except KLdiv, where lower values are preferable, higher values are desirable for all other metrics.

\begin{table}[t]
	\centering
        
	\caption{Performance comparison on IG dataset. \textbf{Bold} and \underline{underline} show the best and second-best performances.}
	\resizebox{0.48\textwidth}{!}{
	\begin{tabular}{lcccccccc}\toprule
        \multirow{2}{*}{} & \multicolumn{4}{c}{\textbf{ZeroIA Setting}} & \multicolumn{4}{c}{\textbf{Fully Supervised Setting}}
		\\ \cmidrule(lr){2-5} \cmidrule(lr){6-9}
		Method & CC↑ & KLdiv↓ & SIM↑ &  AUC↑ & CC↑ & KLdiv↓ & SIM↑ &  AUC↑  \\ \midrule
        ITTI  \cite{itti1998model} & 0.0784  & 4.3553  & 0.0142  & 0.4999 & 0.0754 & 3.1815  & 0.0117 & 0.5301  \\ 
        GBVS \cite{harel2006graph} & 0.0262 & 4.5495  & 0.0234  & 0.5000 & 0.1773 & 2.9498 & 0.0207 & 0.5037  \\ 
        DeepGaze I \cite{KummererTB14} & 0.1873 & 3.0663 & 0.0080 & 0.5570 & 0.1969 & 3.1869 & 0.0081 & 0.6143  \\ 
        DeepGaze IIE \cite{Linardos_2021_ICCV} & 0.2065 & 3.0255 & 0.0091 & 0.5814 & 0.1721 & 3.2357 & 0.0070 & 0.6049  \\ 
        UMB  \cite{Zhu_2023_ICCV} & 0.2121 & 3.9004 & 0.5937 & 0.5326 & 0.2011 & 4.2338 & 0.6147 & 0.5319  \\ 
        MLNet \cite{dodge2018visual} & 0.2937 & 3.8987 & 0.4646 & 0.5972 & 0.3195 & 3.6804 & 0.4153 & 0.5743  \\ 
        ConvNext \cite{liu2022convnet} & 0.2571 & 4.8729 & 0.6110 & 0.5482 & 0.2731  & 4.5154 & 0.5747 & 0.5524  \\ 
        SSwin Transformer \cite{Jiang_2022_CVPR} & 0.2494 & 2.6654 & 0.3309 & 0.5212 & 0.2938 & 3.4309 & 0.3323 & 0.5391  \\ 
        CLIP-base  \cite{radford2021learning} & 0.1403 & 3.6118 & 0.0058 & 0.3516 & 0.1564 & 3.8721 & 0.0074 & 0.3385  \\ 
        CLIP-large  \cite{radford2021learning} & 0.1681 & 3.6498 & 0.0064 & 0.3183 & 0.1690 & 3.8547 & 0.0066 & 0.3356  \\ \midrule 
        IA CLIP-base & \underline{0.4013}  & \textbf{2.7114} & \textbf{0.7205} & \underline{0.5953} & \underline{0.4810}  & \underline{2.2132} & \underline{0.7521}  & \underline{0.6375}   \\ 
        IA CLIP-large & \textbf{0.4488} & \underline{2.8409} & \underline{0.7152} & \textbf{0.6068} & \textbf{0.5106}  & \textbf{2.1409} & \textbf{0.7796}  & \textbf{0.6386}  \\ 

	\bottomrule	
 	\end{tabular}}
  \label{table2}
\end{table}

\subsection{Quantitative Comparison.}
Table \ref{table2} presents the quantitative comparison results in the ZeroIA and the fully supervised settings. We can observe that both versions of our model IA outperform all 10 baselines in both metrics. 
In particular, for CC metric, IA w/ CLIP-L outperforms the third-best MLNet by 52.8\% and 59.8\% on the ZeroIA and the fully supervised settings, respectively.

Moreover, we observe that the manual feature-based ITTI and GBVS approaches fail in the interaction-oriented attention prediction task, which demands higher cognitive capabilities. Additionally, when comparing IA with the similarity map of raw CLIP output, our IA achieves significant improvements across all metrics. For example, IA w/ CLIP-B exceeds raw CLIP-B by 186\% under CC metric of the ZeroIA setting. This suggests that our model effectively activates and utilizes CLIP's knowledge representation capabilities, adapting them to cognitive tasks on interaction.

\subsection{Qualitative Analyses.}
\begin{figure*}[t]
	\centering
	\includegraphics[trim=15 20 12 15, clip=true, width=0.84\textwidth]{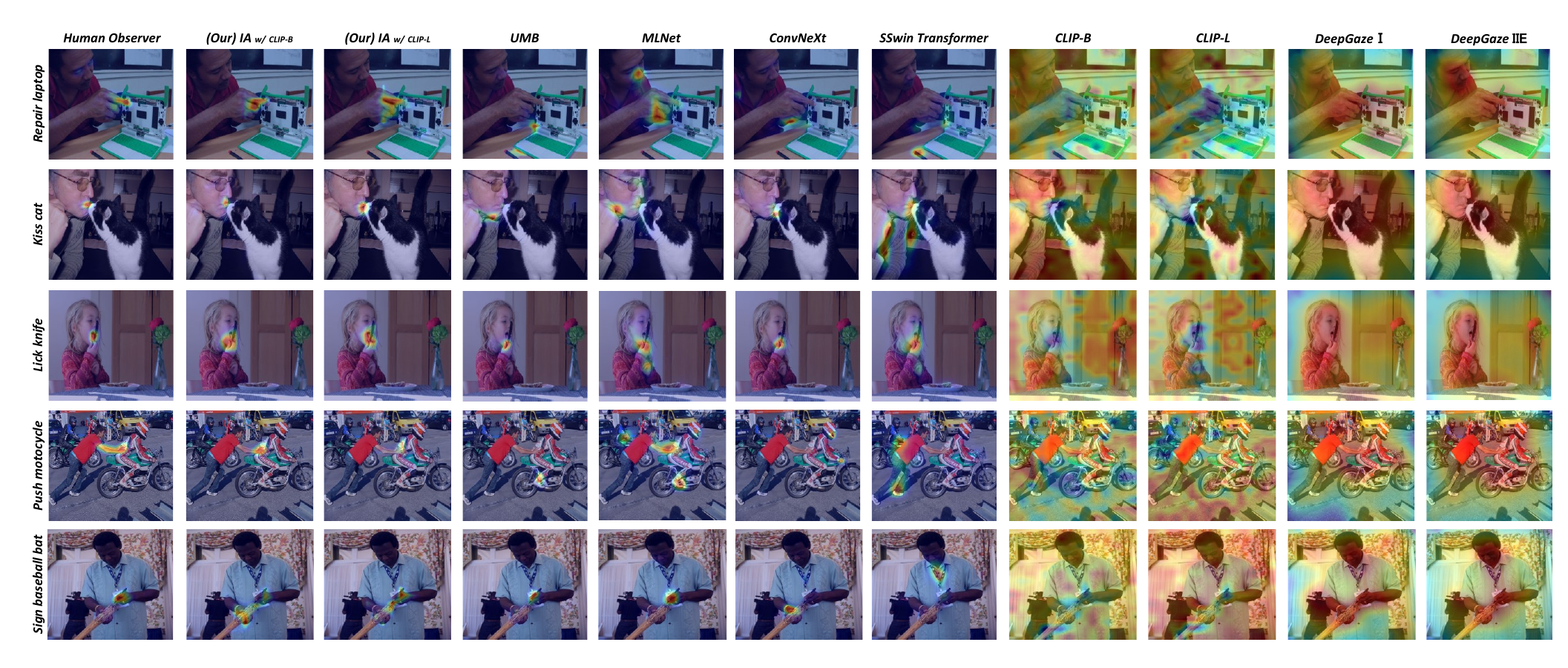}
	\caption{Qualitative comparison of interaction-oriented attention prediction under the ZeroIA setting.}
	\label{fig5}
\end{figure*}

\begin{figure*}[t]
	\centering
	\includegraphics[trim=15 20 20 15, clip=true, width=0.84\textwidth]{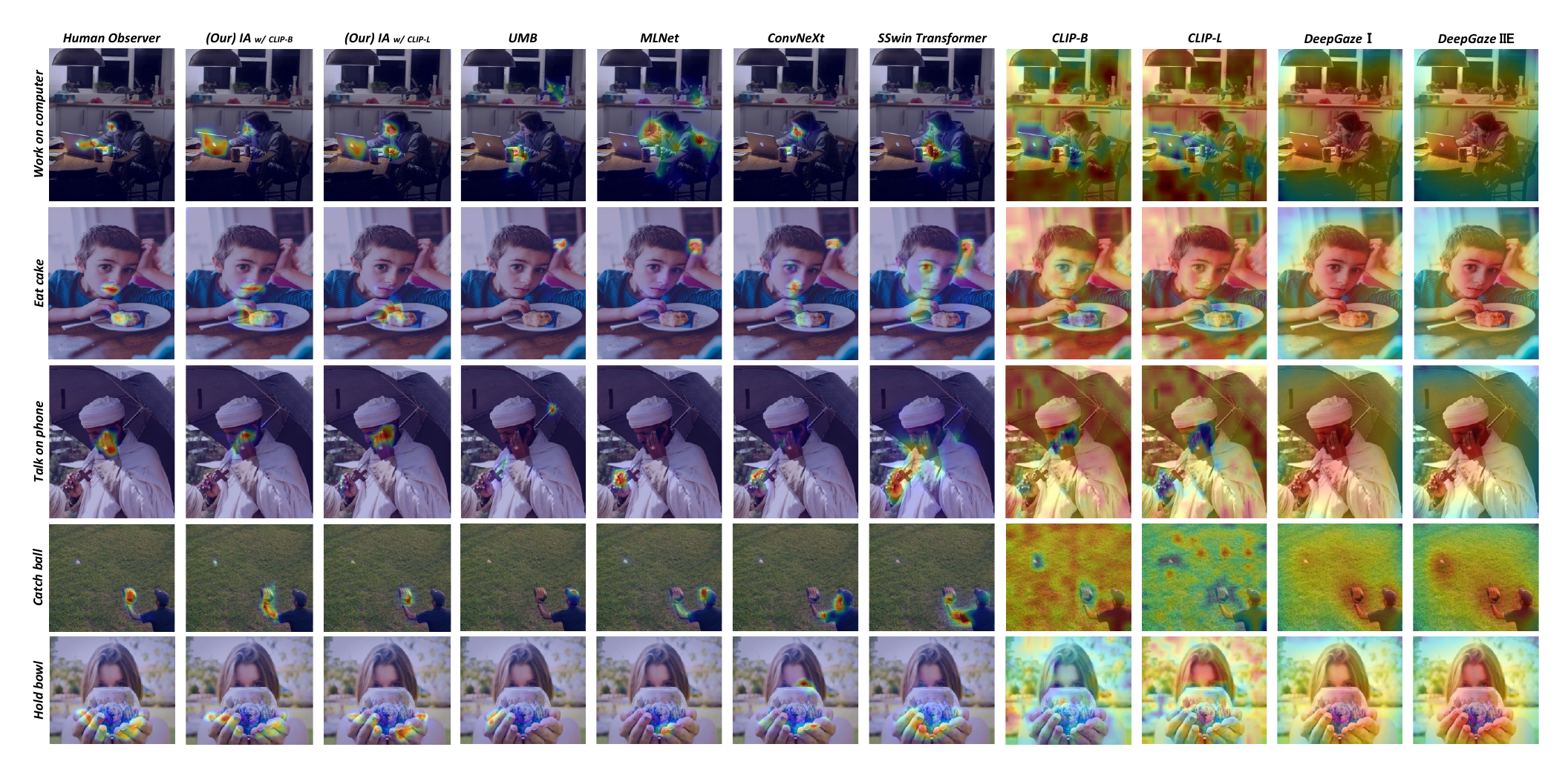}
	\caption{Qualitative comparison of interaction-oriented attention prediction under the fully supervised setting.}
	\label{fig6}
\end{figure*}

As shown in Figures \ref{fig5} and \ref{fig6}, the visualizations of interaction-oriented attention prediction under ZeroIA and fully supervised settings, respectively. Notably, across both settings, IA exhibits attention distributions that are consistently closer to human observers, outperforming the remaining eight methods.

\textbf{Compared with SOTA models.} The previous untrained instance-centric models (columns 6-7) only capture coarse-grained salient persons and objects, but not interaction-related cues. Methods trained on IG prove limited in the complex multi-instance ``push" scenario (Figure \ref{fig5}, row 4) under the ZeroIA setting and also in the ``talk on" scenario (Figure \ref{fig6}, row 3) with a limited number of training samples.

\textbf{Compared with raw CLIP.} IA accurately captures key visual cues, particularly for unseen categories such as ``repair", ``kiss", ``lick", ``push", and ``sign" in Figure \ref{fig5}. Noteworthy is the comparison with columns 5-6, where the similarity maps generated by raw CLIP lack interpretability, displaying properties such as opposite visualization and noisy activation, as highlighted in recent work \cite{li2023clip,li2022exploring}. These observations, along with quantitative results in Table \ref{table2}, underscore the fact that our work effectively activates CLIP's powerful knowledge representation capabilities and improves interpretability.

\subsection{Ablation Study}
To examine the impact of each module in IA, we conduct an ablation study by removing modules in Table \ref{table_ablation}. Specifically, we sequentially exclude the positional adapter (PA), visual adapter (VA), human-object cognitive block (HOCB), and interaction cognitive block (ICB). All ablated models are variations of IA w/ CLIP-B. Results indicate consistent performance degradation across both ZeroIA and fully supervised settings upon removal of any module. Notably, the most significant degradation occurs when the ICB is omitted, emphasizing its crucial role in guiding the learning of interaction-oriented attention by injecting the interaction knowledge. Subsequently, the HOCB, enriched with information about the subjects and objects involved in the interaction, the PA, which incorporates spatial configuration, and the VA, receiving refined visual features, contribute to the model's performance in descending order from high to low.


\begin{table}[t]
	\centering
	\caption{Performance comparison of each modude in IA.}
	\resizebox{0.45\textwidth}{!}{
	\begin{tabular}{lcccccccc}\toprule
        \multirow{2}{*}{} & \multicolumn{4}{c}{\textbf{ZeroIA Setting}} & \multicolumn{4}{c}{\textbf{Fully Supervised Setting}}
		\\ \cmidrule(lr){2-5} \cmidrule(lr){6-9}
		Method & CC↑ & KLdiv↓ & SIM↑ &  AUC↑ & CC↑ & KLdiv↓ & SIM↑ &  AUC↑  \\ \midrule
		
        w/o PA  & 0.3652 & 3.2115 & 0.6733 & 0.5802 & 0.3961 & 2.4676 & 0.6864 & 0.6023  \\ 
        w/o VA  & 0.3838 & 2.9801 & 0.6754 & 0.5832 & 0.4081 & 2.4346 & 0.6893 & 0.6182  \\ 
        w/o HOCB & 0.3498 & 3.3123 & 0.6421 & 0.5743 & 0.3697 & 2.5689 & 0.6484 & 0.6034  \\ 
        w/o ICB  & 0.3249 & 3.6541 & 0.5845 & 0.5644 & 0.3445 & 2.8533 & 0.6023 & 0.5934  \\ \midrule
        IA & 0.4013 & 2.7114 & 0.7205 & 0.5953 & 0.4810  & 2.2132 & 0.7521 & 0.6375 \\ 
	\bottomrule	
 	\end{tabular}}
  \label{table_ablation}
\end{table}

\subsection{Effectiveness of Attention Training Strategy }



\textbf{Aligning human attention enhances HOI model accuracy.} We leverage the VCOCO training subset within the proposed IG, comprising 4,475 samples, approximately 32.4\% of the complete VCOCO training set, as the attention ground truth for additional supervision of existing HOI models. As shown in Table \ref{tabvcoco}, MUREN, UPT, and STIP are improved by 0.6mAP, 0.8mAP, and 0.4mAP, respectively, under the Scenario1 setting. Similarly, under the Scenario2 setting, MUREN, UPT, and STIP are enhanced by 0.7mAP, 1.3mAP, and 0.4mAP, respectively. This highlight the significant potential of aligned human attention in enhancing action comprehension, even with limited training data.

\begin{table}[h]
	\centering
	\caption{Comparison of HOI detection performance on VCOCO.}
	\resizebox{0.39\textwidth}{!}{
		\begin{tabular}{llcc} \toprule
			Method & Backbone & Scenario \#1 &  Scenario \#2   \\  \toprule
			\textbf{\textit{One-stage methods}} &  &  &   \\
			UnionDet \cite{kim2020uniondet} & ResNet-50-FPN & 47.5 & 56.2  \\ 
			IP-Net \cite{wang2020learning} & Hourglass-104 & 51.0 & -  \\ 
			GG-Net \cite{liao2020ppdm} & Hourglass-104 & 54.7 & -  \\ 
			HOTR \cite{kim2021hotr} & ResNet-50 & 55.2 & 64.4  \\ 
			QPIC \cite{tamura2021qpic} & ResNet-101 & 58.3 & 60.7  \\ 
			MUREN \cite{kim2023relational} & ResNet-50 & 66.9 & 69.0  \\ 
			\textbf{MUREN w/ Human} & ResNet-50 & 67.5 & 69.7  \\ 
			\textbf{MUREN w/ IA-B} & ResNet-50 & 67.7 & 70.3  \\ 
			\textbf{MUREN w/ IA-L} & ResNet-50 & 68.0 & 70.4 \\ \midrule
			\textbf{\textit{Two-stage methods}} &  &  &   \\
			VCL \cite{hou2020visual} & ResNet-50 & 48.3 & -  \\ 
			SCG  \cite{zhang2021spatially} & ResNet-50-FPN & 54.2 & -  \\ 
			UPT \cite{zhang2022efficient} & ResNet-50 & 59.8 & 65.5  \\ 
			STIP \cite{zhang2022exploring} & ResNet-50 & 66.0 & 70.7  \\ 
			\textbf{UPT w/ Human} & ResNet-50 & 60.7 & 66.4  \\ 
			\textbf{UPT w/ IA-B} & ResNet-50 & 61.1 & 66.8  \\ 
			\textbf{UPT w/ IA-L} & ResNet-50 & 61.4 & 67.0  \\ 
			\textbf{STIP w/ Human} & ResNet-50 & 66.4 & 71.1  \\ 
			\textbf{STIP w/ IA-B} & ResNet-50 & 67.0 & 71.3  \\ 
			\textbf{STIP w/ IA-L} & ResNet-50 & 67.2 & 71.4 \\ \bottomrule	
		\end{tabular}
	}
	\label{tabvcoco}
\end{table}

\begin{table}[h]
	\centering
	\caption{Comparison of HOI detection performance on HICO-det.}
	\resizebox{0.39\textwidth}{!}{
		\begin{tabular}{llccc} \toprule
			Method & Backbone & Full & Rare &  Non-rare  \\ \toprule
			\textbf{\textit{One-stage methods}} &  &  &  & \\
			UnionDet \cite{kim2020uniondet} & ResNet-50-FPN & 17.58 & 11.72 & 19.33  \\ 
			IP-Net \cite{wang2020learning} & Hourglass-104 & 19.56 & 12.79 & 21.58  \\ 
			PPDM \cite{liao2020ppdm} & Hourglass-104 & 21.73 & 13.78 & 21.40  \\ 
			GG-Net \cite{zhong2021glance} & Hourglass-104 & 23.47 & 16.48 & 25.60  \\ 
			HOTR \cite{kim2021hotr} & ResNet-50 & 25.10 & 17.34 & 27.42  \\ 
			QPIC \cite{tamura2021qpic} & ResNet-101 & 29.90 & 23.92 & 31.69  \\ 
			MSTR \cite{kim2022mstr} & ResNet-50 & 31.17 & 25.31 & 32.92  \\ 
			MUREN \cite{kim2023relational} & ResNet-50 & 32.09 & 27.36 & 33.50  \\ 
			\textbf{MUREN w/ IA-B} & ResNet-50 & 32.45 & 28.90 & 33.51 \\ 
			\textbf{MUREN w/ IA-L} & ResNet-50 & 32.48 & 29.01 & 33.53  \\ \midrule 
			\textbf{\textit{Two-stage methods}} &  &  &  & \\ 
			VCL \cite{hou2020visual} & ResNet-50 & 23.63 & 17.21 & 25.55  \\ 
			ATL \cite{hou2021affordance} & ResNet-50 & 23.67 & 17.64 & 25.47  \\ 
			SCG \cite{zhang2021spatially} & ResNet-50-FPN & 29.26 & 24.61 & 30.65  \\ 
			UPT \cite{zhang2022efficient} & ResNet-50 & 31.93 & 26.70 & 33.49  \\ 
			STIP \cite{zhang2022exploring} & ResNet-50 & 32.22 & 28.15 & 33.43  \\ 
			\textbf{UPT w/ IA-B} & ResNet-50 & 32.20 & 27.59 & 33.58  \\ 
			\textbf{UPT w/ IA-L} & ResNet-50 & 32.47 & 27.76 & 33.72  \\ 
			\textbf{STIP w/ IA-B} & ResNet-50 & 32.32 & 28.47 & 33.59  \\ 
			\textbf{STIP w/ IA-L} & ResNet-50 & 32.64 & 28.90 & 33.76 \\ 
			\bottomrule	
	\end{tabular}
}
\label{tab_hico}
\end{table}

\textbf{Aligning IA-generated attention enhances HOI model accuracy more significantly.} We further enhance the additional supervision of HOI models through large-scale interaction-oriented pseudo-labeling, generated by the proposed IA model on the training sets of VCOCO and HICO-det, respectively.
In Table \ref{tabvcoco}, on the VCOCO dataset, our training strategies consistently obtain performance improvements for MUREN, UPT, and STIP. 
In particular, thanks to the precise emulation of human attention by our proposed IA model, the results of all HOI models reveal that training by a large number of pseudo-attention labels is superior to using limited genuine human attention. On HICO-det dataset, our training strategy similarly achieves consistent performance gains in Table \ref{tab_hico}. Particularly, we observe that our training strategy significantly enhances the performance of the rare set, by 6.0\%, 4.0\%, and 2.7\% on MUREN, UPT, and STIP, respectively. Note that the reported experimental results are all obtained by reproducing these models under the same experimental settings.

\begin{figure}[t]
	\includegraphics[trim=15 20 20 15, clip=true, width=0.44\textwidth]{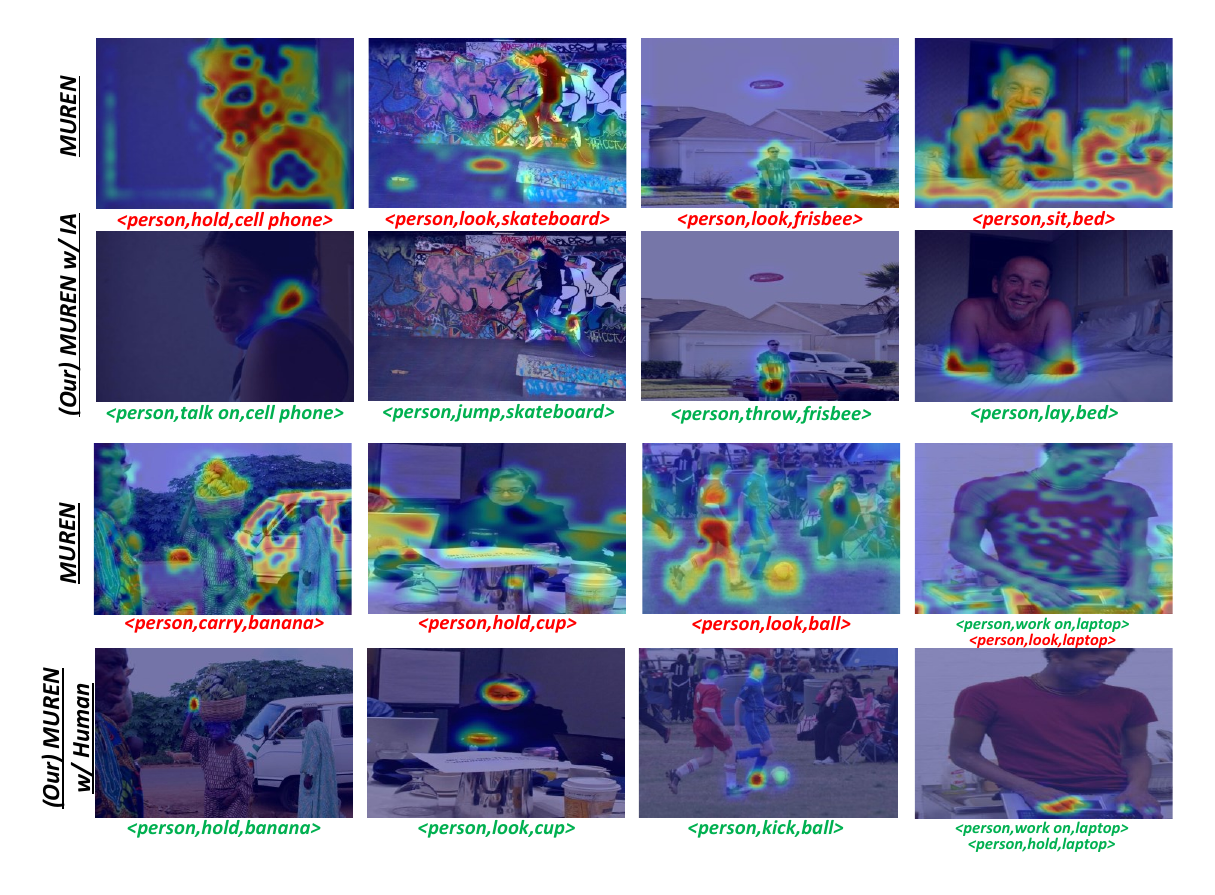}
	\caption{The predicted results and corresponding attention visualizations for MUREN and our (MUREN w/ IA-L, MUREN w/ Human). We mark true positive results in \textcolor[RGB]{0,176,80}{green}, and false positive results in \textcolor{red}{red}. After aligning interaction-oriented attention, the erroneous prediction results are corrected, and the corresponding attention becomes more converged and more interpretable. More in supplementary materials.}
	\label{fig_visualHOI}
\end{figure}

\textbf{Aligning attention makes HOI models more interpretable.}
We present visualizations of the cross-attention maps of interaction branches in original MUREN and the MUREN aligned with interaction-oriented attention, respectively, as shown in Figure \ref{fig_visualHOI}. 
It is evident that the attention map of original MUREN appears fragmented and struggles to focus on interaction-related visual cues, leading to failures in interaction recognition. Conversely, after aligning with interaction-oriented attention, not only are the erroneous results corrected, but the attention map becomes significantly more interpretable and focuses on key regions.

\section{Conclusion}
This paper aims to address a critical gap in interaction-oriented attention by introducing the challenging yet meaningful ZeroIA problem. To achieve this, we collect the IG dataset, marking the first large-scale interaction-centric gaze fixation dataset, featuring 530,000 fixation points across 740 diverse interaction categories.
Subsequently, we present the IA model, ingeniously designed to activate and leverage the knowledge representation capabilities of CLIP, emulating the cognitive processes of human observers to generate high-quality interaction-oriented attention. Extensive experiments demonstrate that the proposed IA model outperforms state-of-the-art attention prediction methods in both zero-shot and fully supervised settings.
Furthermore, we propose an attention-guided HOI training strategy, leveraging interaction-oriented attention to guide existing HOI models for more effective learning. Quantitative and qualitative results demonstrate that both genuine human attention and IA-generated attention significantly enhance the performance and interpretability of existing HOI models, making them more capable of visual reasoning like human observers. This highlights the significant potential of visual attention in action comprehension.

Our endeavor represents the pioneering effort to establish a bi-directional connection between visual attention and HOI detection, starting by directing visual attention through the interaction comprehension task and ending by incorporating this attention to repay existing HOI methods. 
We believe our work can inspire further research on goal-oriented attention and its application in various action-related tasks like classification, VQA, and captioning. Our work also has potential applications in DNNs’ interpretability evaluation, human-AI collaboration, and values alignment research.

\textbf{Acknowledgements.} This work is supported in part by National Natural Science Foundation of China under Grant 62373387, and in part by Natural Science Foundation of Guangdong Province (No. 2023A1515030264).






{\small
\bibliographystyle{ieee_fullname}
\bibliography{egbib}
}

\end{document}